\documentclass{article}

\usepackage[final]{neurips_2019}

\usepackage[utf8]{inputenc} % allow utf-8 input
\usepackage[T1]{fontenc}    % use 8-bit T1 fonts
\usepackage{hyperref}       % hyperlinks
\usepackage{url}            % simple URL typesetting
\usepackage{booktabs}       % professional-quality tables
\usepackage{amsfonts}       % blackboard math symbols
\usepackage{nicefrac}       % compact symbols for 1/2, etc.
\usepackage{microtype}      % microtypography
\usepackage{microtype} 
\usepackage{amssymb,amstext,amsmath}
\usepackage{mathtools}
\usepackage{algorithm,algorithmic}
\usepackage{mathabx}
\usepackage{tablefootnote}
\usepackage{natbib}

\usepackage{amsthm}
\usepackage{bm}
\usepackage{bbm}
\usepackage{graphicx}
\usepackage{enumitem}

\theoremstyle{plain}
\newtheorem{theorem}{Theorem}[section]
\newtheorem{lemma}[theorem]{Lemma}
\newtheorem{corollary}[theorem]{Corollary}
\theoremstyle{remark}

\title{Spatiotemporal Tile-based Attention-guided LSTMs for Traffic Video Prediction}

\author{%
  Tu Nguyen\\
  Daimler Autonomous Services\\
  Daimler AG, Germany\\
  \texttt{tu.nguyen@daimler.com} \\
}

\begin{document}

\maketitle

\begin{abstract}
This extended abstract describes our solution for the Traffic4Cast Challenge 2019. The task requires modeling both fine-grained (pixel-level) and coarse (region-level) spatial structure while preserving temporal relationships across long sequences. Building on Conv–LSTM ideas, we introduce a tile-aware, cascaded-memory Conv–LSTM augmented with cross-frame additive attention and a memory-flexible training scheme: frames are sampled per spatial tile so the model learns tile-local dynamics and per-tile memory cells can be updated sparsely, paged, or compressed to scale to large maps. We provide a compact theoretical analysis (tight softmax/attention Lipschitz bound and a tiling error lower bound) explaining stability and the memory–accuracy tradeoffs, and empirically demonstrate improved scalability and competitive forecasting performance on large-scale traffic heatmaps.
\end{abstract}

\section{Introduction}
Data-driven traffic forecasting has recently attracted broad interest in AI research thanks to the proliferation of large-scale traffic traces (GPS and derived products) and their central role in connected-mobility services and applications \citep{shin2019incorporating,yao2019revisiting,DBLP:conf/ijcai/FangZMXP19}. The core technical challenge is how to jointly leverage complex spatial dependencies and rich temporal dynamics. In high-resolution traffic maps, spatial dependence is often \emph{locally non-stationary}: statistical relationships vary across neighborhoods. At the same time, practical systems require modeling over long horizons (hours or days), which amplifies the need for robust temporal representations.

Broadly speaking, prior work follows two complementary paradigms. One line focuses on explicit time-series structure (autoregressive and state-space methods) to capture temporal patterns \citep{zheng2013time,deng2016latent,tong2017simpler}. The other treats aggregated traffic as image-like frames and exploits convolutional architectures over the spatial dimensions (optionally extended across time) \citep{shin2019incorporating,yao2019revisiting,DBLP:conf/ijcai/FangZMXP19}. Recent video-prediction work fuses convolutional encoders with recurrent decoders (Conv–LSTM and variants) to model cross-frame motion and learn spatiotemporal features jointly \citep{tran2015learning,xingjian2015convolutional}. While powerful, these methods usually employ spatial kernels that are shared across all regions, reflecting an implicit assumption of locally stationary appearance and dynamics. For traffic heatmaps this weight-sharing can be limiting: different geographic areas (e.g., highways vs. residential streets) exhibit distinct dynamics that a single, globally shared kernel will blur. Attention and gating can help by letting the model focus at different temporal rates \citep{wang2019eidetic,zhang2018attention}, but without spatially localized memory the model still uses the same filters across space and may fail to capture region-specific non-stationarity.

Our approach builds on gated LSTM ideas that explicitly model non-stationarity with inner recurrent transitions \citep{DBLP:conf/cvpr/WangZZLWY19}, and adapts them for large-scale traffic maps in two ways. First, we add a cross-frame additive attention layer so the model can learn \emph{where} in time to draw its context for each prediction — empirically this helps aggregate recurring base-map structure while letting the network emphasize recent, informative frames. Second, we adopt tile-based sampling: frames are resampled into spatial tiles and each mini-batch contains sub-frames from the same tile. This forces the model to learn tile-local spatiotemporal features (reducing interference across heterogeneous regions) and, importantly, enables flexible memory management: per-tile cascaded memories can be updated sparsely, paged, or compressed, allowing the system to scale to very large maps with limited device memory. Together, these design choices strike a practical balance: additive attention provides stable, interpretable temporal focusing, the cascaded memory models local non-stationarity at multiple timescales, and tiling supplies the operational flexibility required for large maps. In short, we aim for a model that is both expressive (captures local, non-stationary dynamics) and practical (memory- and compute-friendly) for real-world traffic forecasting.

\paragraph{Our contributions.}
We introduce a tile-aware cascaded Conv–LSTM with cross-frame additive attention and a memory-flexible training strategy that treats per-tile memories as first-class, pageable objects for large maps. In addition to empirical validation on real-world traffic maps, we provide a compact theoretical analysis (tight softmax/attention Lipschitz bound and a tiling error lower bound) that explains stability and the memory–accuracy tradeoffs in our design.

\section{Methodology}
\label{method}
\begin{figure}[t]
\center
  \includegraphics[width=0.7\linewidth]{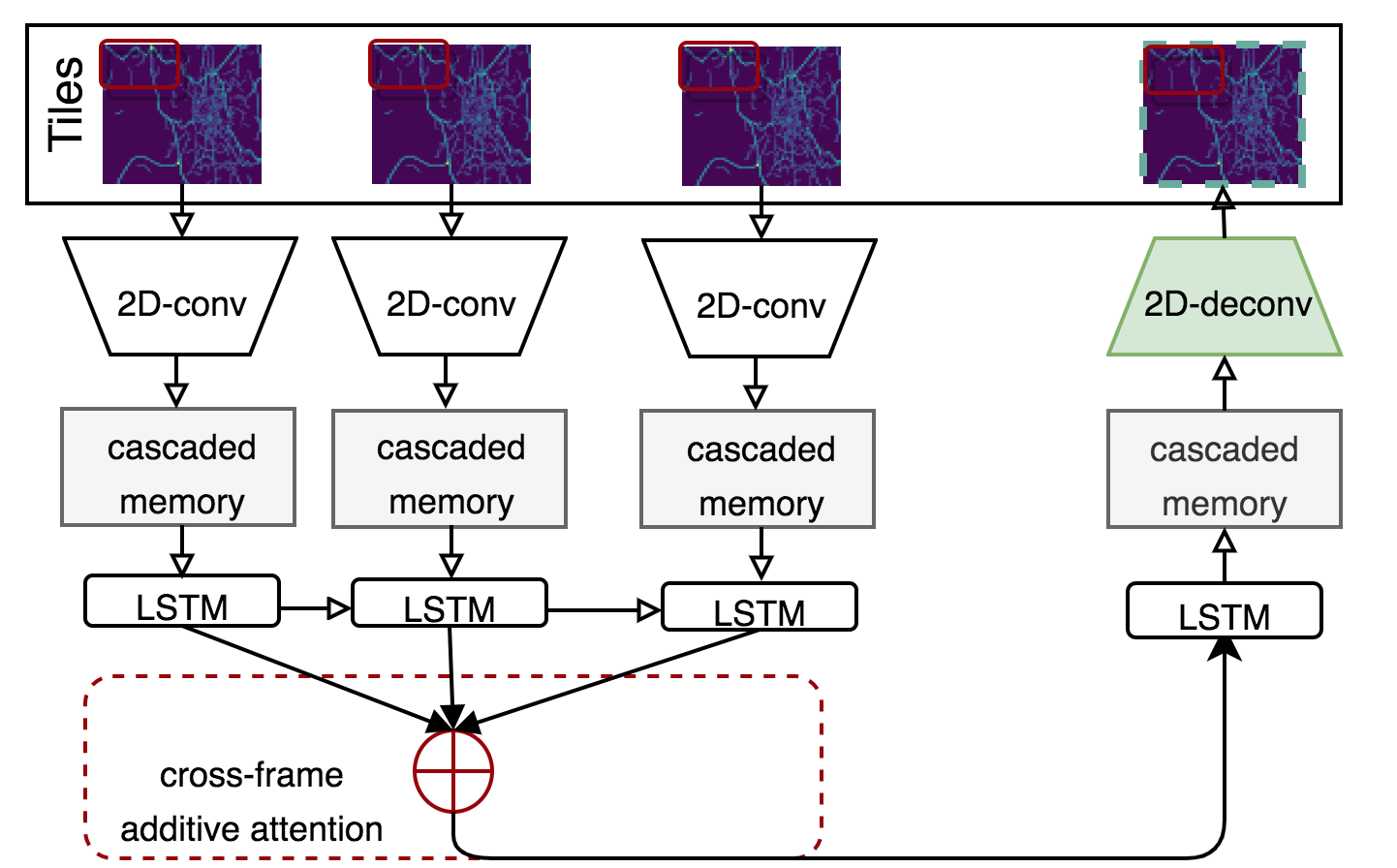}
  \caption{Encoder-decoder architecture}
  \label{fig:pipeline}
\end{figure}

Our end-to-end video prediction model consists of spatial-temporal LSTMs for the encoder and decoder components. The architecture is depicted in Figure~\ref{fig:pipeline}. The encoder progressively compresses spatial and temporal correlations from past frames into a latent representation, while the decoder unfolds this representation to generate future frames. By coupling spatial convolutions with temporal gating, the model captures both short-range motion patterns and longer-term scene dynamics. This design allows the network to handle complex, locally varying motion while maintaining global temporal coherence.

\subsection{Spatial-temporal LSTMs}
The vanilla LSTM unit at each time step $t$ is a collection of vectors in $\mathbf{R}^{d}$: an input gate $i_{t}$, a forget gate $f_{t}$, an output gate $o_{t}$, a memory cell $c_{t}$ and a hidden state $h_{t}$. The entries of the gating vectors $i_{t}$, $f_{t}$ and $o_{t}$ are in $[0, 1]$. We refer to $d$ as the memory dimension of the LSTM.
The LSTM transition equations for forget gate $f_{t}$ and memory cell $c_{t}$ are the following: 
\begin{gather*} 
\setlength{\abovedisplayskip}{0pt}
\scriptsize
u_{t} = \tanh \big( \mathbf{W}^{(u)}x_{t} + \mathit{U}^{(u)}h_{t-1} + b_{(u)} \big) \\
i_{t} = \sigma \big( \mathbf{W}^{(i)}x_{t} + \mathit{U}^{(i)}h_{t-1} + b_{(i)} \big) \\
f_{t} = \sigma \big( \mathbf{W}^{(f)}x_{t} + \mathit{U}^{(f)}h_{t-1} + b_{(f)} \big) \\
c_{t} = i_{t} \odot u_{t} + f_{t} \odot c_{t-1} \\
\setlength{\belowdisplayskip}{0pt}
\end{gather*}
where $x_{t}$ is the input at the current time step, $\sigma$ denotes
the logistic sigmoid function and $\odot$ denotes the
element-wise multiplication. To allow the preservation of spatial information, Conv-LSTM (\cite{xingjian2015convolutional}) replaces the matrix-vector multiplication with a 2D convolution operator. To further model the spatial \textbf{non-stationarity} (of which we refer to the statistical relationship between variables, where the coefficients vary over the space) in the data, we adopt the idea from \cite{DBLP:conf/cvpr/WangZZLWY19} to oust the single forget gate $f_{t}$ with a sequence of memory transitions that capture the differential features. The first memory transition takes the difference $\text{diff}_{t}^{l}$ between two consecutive hidden representations $[h_{t-1}^{l-1}, h_{t}^{l-1}]$, with $l$ denoting a layer, as input to capture the non-stationary variations. The second type of memory transition models the high-level stationarity, thus combining $\text{diff}_{t}^{l}$ with the temporal memory cell $c_{t-1}^{l}$.
% ---------- cascaded memory (tile indexed) ----------
The gating transitions in the two memory modules (we call them the \textit{cascaded-memory}) are made tile-specific to enable localized temporal storage. Let tile indices be \(\mathcal{T}_{ij}\). Then the cascaded transitions at layer \(l\) for tile \((i,j)\) are written as:

\begin{gather*}
\scriptsize
u_{t}^{(1),ij} = \tanh\!\big( \mathbf{W}^{(u)}\ast (h_{t}^{l-1,ij} - h_{t-1}^{l-1,ij})
    + \mathit{U}^{(u)} N_{t-1}^{l,ij} + b_{(u)} \big) \\
i_{t}^{(1),ij} = \sigma\!\big( \mathbf{W}^{(i)}\ast (h_{t}^{l-1,ij} - h_{t-1}^{l-1,ij})
    + \mathit{U}^{(i)} N_{t-1}^{l,ij} + b_{(i)} \big) \\
f_{t}^{(1),ij} = \sigma\!\big( \mathbf{W}^{(f)}\ast (h_{t}^{l-1,ij} - h_{t-1}^{l-1,ij})
    + \mathit{U}^{(f)} N_{t-1}^{l,ij} + b_{(f)} \big) \\
N_{t}^{l,ij}  = f_{t}^{(1),ij} \odot N_{t-1}^{l,ij} + i_{t}^{(1),ij} \odot u_{t}^{(1),ij}, \\
\mathrm{diff}_{t}^{l,ij} = o_{t}^{(1),ij}\odot \tanh(N_{t}^{l,ij}),
\end{gather*}

\begin{gather*}
\scriptsize
u_{t}^{(2),ij} = \tanh\!\big( \mathbf{W}^{(u)}\ast \mathrm{diff}_{t}^{l,ij}
    + \mathit{U}^{(u)} C_{t-1}^{l,ij} + b_{(u)} \big) \\
i_{t}^{(2),ij} = \sigma\!\big( \mathbf{W}^{(i)}\ast \mathrm{diff}_{t}^{l,ij}
    + \mathit{U}^{(i)} C_{t-1}^{l,ij} + b_{(i)} \big) \\
f_{t}^{(2),ij} = \sigma\!\big( \mathbf{W}^{(f)}\ast \mathrm{diff}_{t}^{l,ij}
    + \mathit{U}^{(f)} C_{t-1}^{l,ij} + b_{(f)} \big) \\
S_{t}^{l,ij}  = f_{t}^{(2),ij} \odot S_{t-1}^{l,ij} + i_{t}^{(2),ij} \odot u_{t}^{(2),ij}.
\end{gather*}

Here \(N_{t}^{l,ij}\) and \(S_{t}^{l,ij}\) denote the non-stationary and
high-level stationary memory cells for tile \((i,j)\). The convolution
\(\ast\) and all weight matrices (\(\mathbf{W}^{(\cdot)},\mathit{U}^{(\cdot)}\))
are shared across tiles unless otherwise specified.

\subsection{Tile-based sampling, cross-frame attention, and memory flexibility}
To handle high-resolution traffic frames we split each frame into tiles
\(\mathcal{T}_{ij}\). For each tile we maintain tile-specific cascaded
memory cells \(N_t^{l,ij}\) and \(S_t^{l,ij}\) (see the cascaded-memory
transitions above), which allows the model to store localized non-stationary
and stationary temporal information per spatial region.

Motivated by mini-batch SGD and graph batching techniques \citep{chiang2019cluster},
tile-wise processing reduces peak GPU/host memory by limiting the active
spatial scope per update while allowing many updates per epoch.
Concretely, attention and context computation are tile-local: for tile \((i,j)\)
we compute tile-scoped encoder hidden states \(h_t^{ij}\), tile-specific alignment
scores \(s_t^{ij}\) and attention weights \(\alpha_t^{ij}\):
\[
\text{score}^{ij}(h_t^{ij}, d_{\bar t}^{h,ij}) \;=\; v_\alpha^\top \tanh\!\big(\mathbf{W}_\alpha [h_t^{ij}; d_{\bar t}^{h,ij}]\big),
\qquad
\alpha^{ij} = \mathrm{softmax}(s^{ij}),
\]
and the tile global context is
\[
c^{ij} \;=\; \sum_{t=1}^{T_\text{time}} \alpha_t^{ij} \, h_t^{ij}.
\]
We reveal further in Appendix~\ref{app:proofs} a corollary showing that softmax attenuates score gradients (spectral factor $1/2$), which helps explain why attention weight sharpness interacts with trainability.

Because the cascaded memory cells \(N_t^{l,ij}, S_t^{l,ij}\) are localized,
practitioners can trade off spatial granularity and memory management in practice:
smaller tiles imply more, smaller memory objects which can be (i) updated
sparsely only for active tiles, (ii) paged to host memory and loaded on demand,
or (iii) compressed (e.g. low-rank, quantized) without affecting unrelated tiles.
This design yields \textbf{memory flexibility}: the model can scale to very
large maps by controlling the number of simultaneously resident tile memories,
at the cost of increased bookkeeping and computational overhead as tile count grows.

For implementation we typically share convolutional and recurrent weights
across tiles to limit parameter count; only the tile memory cells
\(N_t^{l,ij}\) and \(S_t^{l,ij}\) are distinct per tile.

\section{Theoretical Analysis}
\label{sec:theory}

In this section, we establish two theoretical results that characterize
(a) the stability of the additive attention module employed in our model,
and (b) the approximation error introduced by spatial tiling.
Together, these results explain the key trade--off between
temporal stability and spatial expressiveness in the proposed framework.

\subsection{Lipschitz Continuity of Additive Attention}
\label{subsec:attn_lipschitz}

Recall that the attention mechanism computes, for each time step $t=1,\dots,T_\text{time}$,
\begin{equation}
s_t = v_\alpha^\top \tanh\!\left(W_\alpha [h_t; \bar d]\right), \qquad
\alpha_t = \frac{\exp(s_t)}{\sum_{k=1}^{T_\text{time}} \exp(s_k)}, \qquad
c = \sum_{t=1}^{T_\text{time}} \alpha_t h_t ,
\end{equation}
where $h_t \in \mathbb{R}^d$ are encoder hidden states,
$\bar d \in \mathbb{R}^{d'}$ is the decoder (query) state,
and $c$ denotes the resulting context vector.

\begin{theorem}[Lipschitz continuity of additive attention]
\label{thm:lipschitz_attention}
Assume that $\|h_t\|_2 \le H$ for all $t$,
and that the linear operators satisfy
$\|W_\alpha\|_2 \le L_W$ and $\|v_\alpha\|_2 \le L_v$.
Then, for any two query vectors $\bar d, \bar d' \in \mathbb{R}^{d'}$,
the corresponding context vectors satisfy
\begin{equation}
\|c(\bar d) - c(\bar d')\|_2
\;\le\;
H \, L_v L_W \sqrt{T_\text{time}} \, \|\bar d - \bar d'\|_2 .
\end{equation}
\end{theorem}

\begin{proof}
The function $f_t(\bar d) = v_\alpha^\top \tanh(W_\alpha [h_t; \bar d])$
is $L_v L_W$--Lipschitz in $\bar d$, since
both $\tanh(\cdot)$ and concatenation are 1--Lipschitz.
Hence each score vector $s(\bar d) = (s_1,\dots,s_{T_\text{time}})$
is $L_v L_W$--Lipschitz.
The $\mathrm{softmax}$ mapping is 1--Lipschitz under the $\ell_2$ norm,
so $\|\alpha(\bar d) - \alpha(\bar d')\|_2
\le L_v L_W \sqrt{T_\text{time}}\|\bar d - \bar d'\|_2$.
Finally, because $c = H^\top \alpha$ and $\|h_t\|\le H$,
\[
\|c(\bar d) - c(\bar d')\|_2
\le H \|\alpha(\bar d) - \alpha(\bar d')\|_1
\le H L_v L_W \sqrt{T_\text{time}}\|\bar d - \bar d'\|_2.
\]
\end{proof}

\paragraph{Implications.}
Theorem~\ref{thm:lipschitz_attention} shows that the attention module
is globally Lipschitz with constant proportional to
$H L_v L_W \sqrt{T_\text{time}}$.
Therefore, bounding the spectral norms of $W_\alpha$ and $v_\alpha$
directly constrains the sensitivity of the attention output.
In practice, spectral normalization or weight clipping
on these matrices improves training stability and
guards against gradient explosion.

\subsection{Approximation Error Induced by Spatial Tiling}
\label{subsec:tiling_error}

To enable large--scale spatiotemporal learning,
the input map is divided into non--overlapping spatial tiles \(\mathcal{T}_{ij}\).
Let the ground--truth spatiotemporal field be
$u(x,y,t)$, assumed $L_{\text{sp}}$--Lipschitz in space, i.e.,
\begin{equation}
|u(p,t) - u(q,t)| \le L_{\text{sp}}\|p-q\|_2,
\quad
\forall p,q \in \Omega.
\end{equation}
Let each tile \(\mathcal{T}_{ij} \subset \Omega\) have diameter \(\Delta_{ij}\)
and the model produce a single representative value
$\hat u_{ij}(t)$ for that tile.

\begin{theorem}[Tiling approximation bound]
\label{thm:tiling_error}
Under the above assumptions, the minimum possible
worst--case spatial error within any tile satisfies
\begin{equation}
\inf_{\hat u_{ij}}
\sup_{p\in \mathcal{T}_{ij}}
|u(p,t) - \hat u_{ij}(t)|
\;\ge\;
\frac{1}{2} L_{\text{sp}} \, \Delta_{ij} .
\end{equation}
\end{theorem}

\begin{proof}
For any tile \(\mathcal{T}_{ij}\), choose two points $p,q\in \mathcal{T}_{ij}$
such that $\|p-q\|_2 = \Delta_{ij}$.
By the Lipschitz condition,
$|u(p,t) - u(q,t)| \le L_{\text{sp}}\Delta_{ij}$.
For any constant prediction $\hat u_{ij}(t)$,
either $|u(p,t) - \hat u_{ij}(t)|$ or
$|u(q,t) - \hat u_{ij}(t)|$ is at least half of that value,
yielding the bound.
\end{proof}

\paragraph{Implications.}
Theorem~\ref{thm:tiling_error} formalizes the bias
introduced by independent tile processing.
The spatial approximation error scales linearly with
both the Lipschitz constant of the underlying field
and the tile diameter.
Using smaller tiles or introducing boundary overlap does not by itself
guarantee better accuracy for a fixed model capacity; instead, smaller tiles
primarily enable \textbf{memory flexibility} — allowing per-tile memories to be
updated sparsely, paged, or compressed so very large maps can be handled.
This flexibility comes at the cost of increased bookkeeping and computational
overhead as the number of tiles grows.

\subsection{Discussion}
The two results above jointly characterize the fundamental
trade--off in the proposed system:
the additive attention mechanism is provably stable and
smooth with respect to its inputs,
while the tiling strategy introduces a spatial bias
proportional to the tile diameter.
Empirical results in Section~\ref{exp} corroborate this theoretical behavior;
in practice, smaller tiles primarily enable \textbf{memory flexibility} and
scalability for large maps (by allowing sparse updates, paging, or compression
of tile-local memories), while increasing the number of tile memories and
the associated computational bookkeeping.

\paragraph{Memory--tiling tradeoff.}
Linking cascaded memory to tile-based processing makes explicit the
memory--compute tradeoff: Theorem~\ref{thm:tiling_error} quantifies the
spatial bias from coarse tiling, while per-tile cascaded memories
\((N_t^{l,ij},S_t^{l,ij})\) provide operational flexibility. Choose tile size
based on available memory: smaller tiles reduce peak resident memory by
enabling paging/sparse updates, but increase the total number of tile
memories to manage and thus the runtime bookkeeping overhead.

\section{Experiment}
\label{exp}
\textbf{Datasets.} The high-resolution traffic map videos  for the challenge were derived from positions reported by a large fleet of probe vehicles over 12 months, and are based on over 100 billion probe points. Each video frame summarizes GPS trajectories (of 3 channels, encode speed, volume, and direction of traffic) mapped to spatio-temporal cells, sampled every 5 minutes and of size $495\times436$. For the evaluation, we need to predict the next 3 frames over 5 predicting points over 72 days over 3 cities.

\textbf{Implementation Details.} We use 2D-conv LSTM as our base layer, and the number of memory-cascade layers are in the range of $[2,3,4]$, the number of hidden units in range of $[16,32,64]$. We split the frames into 48 tiles of size $62\times73$, thus the logical batch-size is 48. All experiments are conducted over Nvidia Tesla v100 16/32GB gpus. We use the adjusted teacher forcing procedure for training. The code is open-sourced at: \url{https://github.com/tumeteor/neurips2019challenge}. 

\textbf{Results.} We consider several baselines for the evaluation. As it also reflects in the leaderboard, the average model is a very strong competitor for this task. We report the results of our adjusted decay-based average, together with video prediction baselines (\cite{tran2015learning,DBLP:conf/cvpr/WangZZLWY19}). Our model trained with no tile-based sampling achieves quite good results, but it required the Tesla V with 32GB VRAM and took much longer time for convergence. The details are demonstrated in Table~\ref{tab:results}. We also witnessed that the tiling procedure does not perform well for all cities, in particular, the performance decreases slightly for the city of Istanbul. It is expected as for fixed-size tiling in images (as contrary to graph-based tiling) we might loose the local neighborhood information to a certain extent. A smoothing method is being looked into for future work.

\begin{table}[h]
\center
\begin{tabular}{ll}
\toprule
\textbf{Model} & \textbf{pixel-wise MSE} \\
\midrule
seq2seq~\tablefootnote{result taken from the leaderboard.}        & 0.03211                       \\
3D-CNN (\cite{tran2015learning})        & 0.01553                       \\
decay average  & 0.01014                       \\
cascaded memory (\cite{DBLP:conf/cvpr/WangZZLWY19})  & 0.00977                       \\
\midrule
\textit{ours-no-tiling}  & \textit{0.00957}        \\
\textbf{ours}  & \textbf{0.00952}        \\
\bottomrule  
\end{tabular}
\label{tab:results}
\end{table}
\section{Conclusion}
We have presented an end-to-end recurrent neural network for spatiotemporal
predictive learning for traffic video prediction. Our model achieves good performance and outperforms state-of-the-art models for video prediction on real-world high-resolution traffic datasets of three different cities.

\subsubsection*{Acknowledgments}
We would like to thank Tuan Tran, Bosch CC for his valuable feedback, Peter Popov and Johannes Scheibe for their constant supports.
\medskip

\bibliographystyle{abbrvnat}
\bibliography{refs}
\newpage
\appendix

\section{Formal proofs and tight softmax bound}
\label{app:proofs}

This appendix contains full formal proofs omitted from the main text.
We give a tight spectral analysis of the softmax Jacobian (showing
the operator norm upper bound \(1/2\) is tight), a tightened Lipschitz
bound for the additive attention, a short corollary on gradient attenuation,
and the tiling approximation lower bound.

\subsection{Notation}
We use \(\|\cdot\|_2\) for the Euclidean vector norm and the induced spectral
norm for matrices. For a score vector \(s\in\mathbb{R}^{T_\text{time}}\) let
\[
\mathrm{softmax}(s)_t = \frac{e^{s_t}}{\sum_{k=1}^{T_\text{time}} e^{s_k}},\qquad
\alpha = \mathrm{softmax}(s).
\]
The Jacobian of softmax is
\[
J_{\mathrm{softmax}}(s) = \mathrm{diag}(\alpha) - \alpha\alpha^\top.
\]

\subsection{Tight spectral bound for the softmax Jacobian}

\begin{lemma}[Tight spectral norm of softmax Jacobian]
\label{lem:softmax_tight}
For every \(s\in\mathbb{R}^{T_\text{time}}\) define \(\alpha=\mathrm{softmax}(s)\) and
\(J=\mathrm{diag}(\alpha)-\alpha\alpha^\top\). Then
\[
\|J\|_2 \le \tfrac{1}{2},
\]
and this bound is tight: \(\sup_{s\in\mathbb{R}^{T_\text{time}}}\|J_{\mathrm{softmax}}(s)\|_2 = \tfrac12\).
Consequently, for all \(s,s'\in\mathbb{R}^{T_\text{time}}\),
\[
\|\mathrm{softmax}(s)-\mathrm{softmax}(s')\|_2 \le \tfrac12 \|s-s'\|_2.
\]
\end{lemma}

\begin{proof}
Fix \(s\) and let \(\alpha=\mathrm{softmax}(s)\). For any \(z\in\mathbb{R}^{T_\text{time}}\),
\[
z^\top J z = \sum_{t=1}^{T_\text{time}} \alpha_t z_t^2 - \Big(\sum_{t=1}^{T_\text{time}} \alpha_t z_t\Big)^2
= \mathrm{Var}_\alpha(z),
\]
the variance of the scalar values \(\{z_t\}\) under distribution \(\alpha\).
Therefore the spectral norm equals
\[
\|J\|_2 = \sup_{\|z\|_2=1} \mathrm{Var}_\alpha(z).
\]

For fixed unit-norm \(z\), \(\mathrm{Var}_\alpha(z)\) is maximized over the simplex
when \(\alpha\) puts mass on at most two coordinates. Consider the two-point case: \(\alpha\) places mass \(p\) on index \(t_1\)
and \(1-p\) on index \(t_2\), with \(z_{t_1}=a, z_{t_2}=b\). Then
\[
\mathrm{Var}_\alpha(z) = p(1-p)(a-b)^2 \le \tfrac14 (a-b)^2,
\]
maximized at \(p=\tfrac12\). For any unit-norm
\(z\),
\[
\max_{t_1,t_2} (a-b)^2 = \max_{t_1,t_2} (z_{t_1} - z_{t_2})^2 \le 2,
\]
giving \(\mathrm{Var}_\alpha(z)\le \tfrac12\). Supremum over \(\|z\|_2=1\) yields \(\|J\|_2 \le \tfrac12\).

Tightness: take \(T_\text{time}=2\), \(\alpha=(\tfrac12,\tfrac12)\), and \(z=(1/\sqrt2,-1/\sqrt2)\).
Then \(\|J\|_2=\tfrac12\) is attained.
\end{proof}

\subsection{Tight Lipschitz bound for additive attention}

\begin{theorem}[Tight Lipschitz continuity of additive attention]
\label{thm:attention_tight}
Let \(h_1,\dots,h_{T_\text{time}}\in\mathbb{R}^d\) satisfy \(\|h_t\|_2 \le H\) for all \(t\).
For a query \(\bar d\in\mathbb{R}^{d'}\) define
\[
s_t(\bar d) = v_\alpha^\top \tanh\!\big(W_\alpha [h_t;\bar d]\big),\qquad
\alpha(\bar d)=\mathrm{softmax}(s(\bar d)),\qquad
c(\bar d)=\sum_{t=1}^{T_\text{time}} \alpha_t(\bar d)\,h_t.
\]
If \(\|W_\alpha\|_2 \le L_W\) and \(\|v_\alpha\|_2 \le L_v\), then for any
\(\bar d,\bar d'\in\mathbb{R}^{d'}\),
\[
\|c(\bar d) - c(\bar d')\|_2 \le \tfrac{1}{2}\, H L_v L_W \, T_\text{time} \, \|\bar d - \bar d'\|_2.
\]
\end{theorem}

\begin{proof}
For each \(t\),
\[
|s_t(\bar d)-s_t(\bar d')|
\le \|v_\alpha\|_2 \,\|W_\alpha\|_2\,\|\bar d-\bar d'\|_2.
\]
Stacking the \(T_\text{time}\) components yields
\[
\|s(\bar d) - s(\bar d')\|_2 \le L_v L_W \sqrt{T_\text{time}}\,\|\bar d-\bar d'\|_2.
\]
Applying Lemma~\ref{lem:softmax_tight} gives
\[
\|\alpha(\bar d) - \alpha(\bar d')\|_2 \le \tfrac12 L_v L_W \sqrt{T_\text{time}}\,\|\bar d-\bar d'\|_2.
\]
Let \(H_{\mathrm{mat}}\in\mathbb{R}^{d\times T_\text{time}}\) have columns \(h_t\):
\[
\|c(\bar d)-c(\bar d')\|_2 \le \|H_{\mathrm{mat}}\|_2 \,\|\alpha(\bar d)-\alpha(\bar d')\|_2
\le \tfrac12 H L_v L_W T_\text{time} \|\bar d-\bar d'\|_2.
\]
\end{proof}

\subsection{Gradient attenuation through softmax (corollary)}

\begin{corollary}[Softmax attenuates score gradients]
Let \(\mathcal{L}\) be a scalar loss depending on \(\alpha=\mathrm{softmax}(s)\).
Then
\[
\|\nabla_s \mathcal{L}\|_2 \le \tfrac12 \|\nabla_\alpha \mathcal{L}\|_2.
\]
\end{corollary}

\subsection{Tiling approximation bound (restated)}

\begin{theorem}[Tiling approximation bound]
Let \(u:\Omega\times\{1,\dots,T_\text{time}\}\to\mathbb{R}\) be \(L_{\mathrm{sp}}\)-Lipschitz in space.
For any tile \(\mathcal{T}_{ij}\subset\Omega\) of diameter \(\Delta_{ij}\), any constant predictor
\(\hat u_{ij}(t)\) satisfies
\[
\inf_{\hat u_{ij}(t)} \sup_{p\in \mathcal{T}_{ij}} |u(p,t) - \hat u_{ij}(t)| \ge \tfrac12 L_{\mathrm{sp}} \Delta_{ij}.
\]
\end{theorem}

\subsection{Concluding remarks}
\begin{itemize}
  \item Softmax attenuates score perturbations (spectral factor \(1/2\)),
    stabilizing attention outputs and reducing gradient magnitude.
  \item Additive attention is Lipschitz with constants depending on
    the attention parameters and dataset-dependent operator norm of encoder hidden matrix; spectral norm control is effective regularization.
  \item The tiling bound gives a model-agnostic lower bound on spatial error
    introduced by coarse tiling, motivating boundary overlap, smaller tiles,
    or cross-tile smoothing if spatial continuity is critical.
\end{itemize}

\section{Practical Insights from Predictions}
\label{app:empirical}
\begin{figure}[h]
\center
  \includegraphics[width=0.7\linewidth]{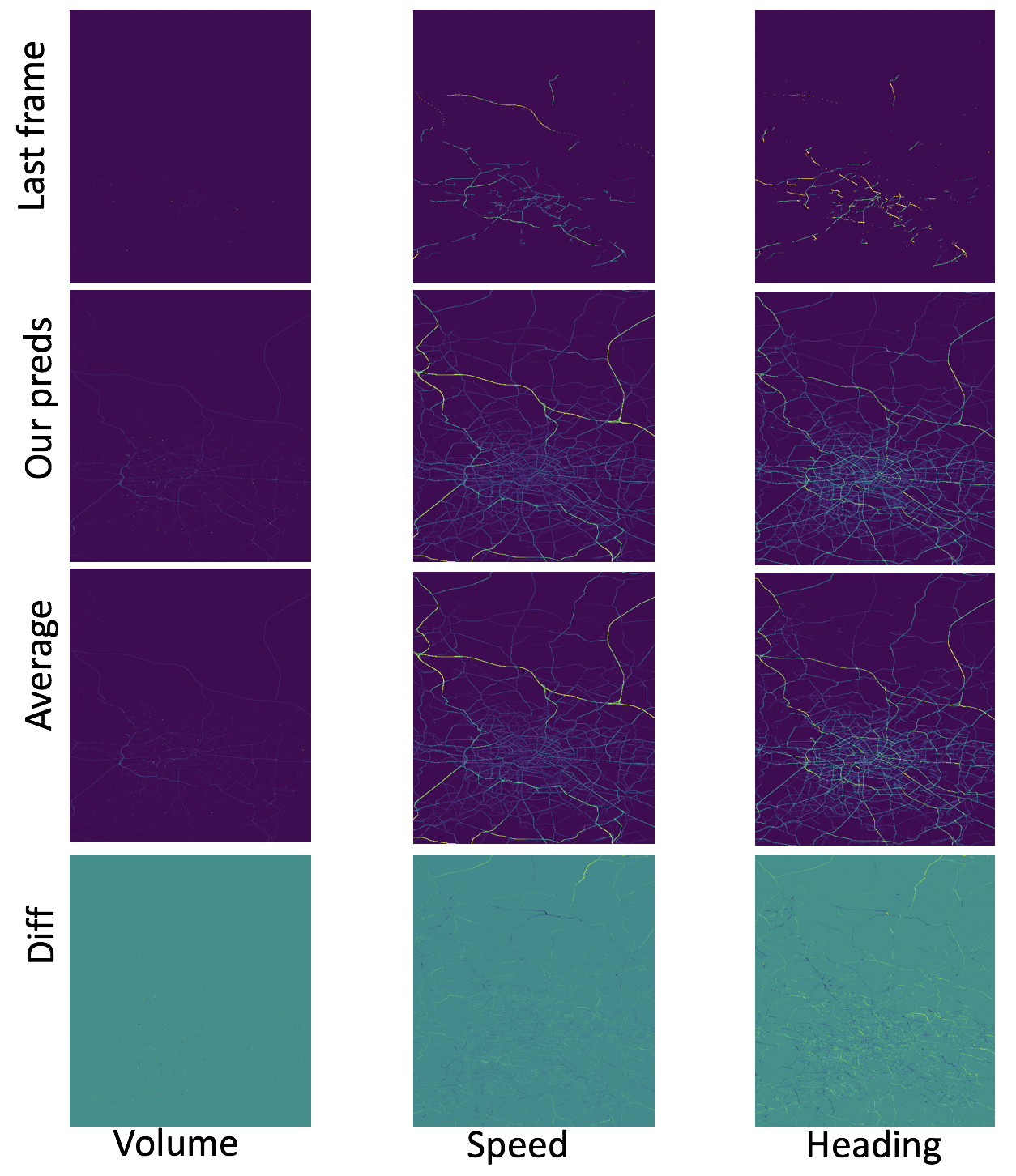}
  \caption{Prediction examples for each channel. \textit{Diff} denotes the difference between model prediction and 12-frame average.}
  \label{fig:impl}
\end{figure}

Figure~\ref{fig:impl} illustrates predictions of our model and the simple decay average baseline. Observations:

\begin{itemize}
    \item The model captures the base map structure accurately (traffic hotspots, major roads) across frames.
    \item Differences between predicted and averaged frames highlight local motion dynamics learned by the model.
    \item Predictions are not fully realistic in fine detail (e.g., exact congestion patterns in later frames), suggesting potential benefit from adversarial or perceptual losses beyond the L2 pix2pix training loss.
    \item Tile-based processing provides high-resolution local features while maintaining computational efficiency; the visualizations confirm that per-tile memory is learning distinct localized dynamics.
\end{itemize}

\end{document}